\pdfoutput=1
\documentclass[11pt]{article}

\usepackage[]{ACL2023}

\usepackage{times}
\usepackage{latexsym}

\usepackage[T1]{fontenc}
\usepackage[utf8]{inputenc}

\usepackage{microtype}

\usepackage{inconsolata}
\usepackage{graphicx}
\usepackage{hyperref}
\usepackage{url}
\usepackage{graphicx}%
\usepackage{bm}
\usepackage{multirow}%
\usepackage{multicol}
\usepackage{amsmath,amssymb,amsfonts}%
\usepackage{amsthm}%
\usepackage{mathrsfs}%
\usepackage{xcolor}%
\usepackage{textcomp}%
\usepackage{manyfoot}%
\usepackage{booktabs}%
\usepackage{algorithm}
\usepackage{algorithmic}
\usepackage{listings}%
\usepackage{appendix}
\usepackage{tabularx}
\begin{document}

% \title{\hspace{-0.4cm}EmoBench: Evaluating the Emotional Intelligence of Large Language Models}
\title{COMET\includegraphics[height=2em]{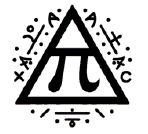}: ``Cone of experience'' enhanced\\large multimodal model for mathematical problem generation}

% \title{COMET: ``Cone of experience'' enhanced large multimodal model for mathematical problem generation}

\author{
   \textbf{Sannyuya Liu}$^{1,2}$,
   ~~\textbf{Jintian Feng}$^{1,2}$,
   ~~\textbf{Zongkai Yang}$^{1,2}$,
   ~~\textbf{Yawei Luo}$^{3}$,\\
   ~\textbf{Qian Wan}$^{1,2*}$,
   ~~\textbf{Xiaoxuan Shen}$^{1,2*}$,
   ~~\textbf{Jianwen Sun}$^{1,2*}$ \\
   $^1$National Engineering Research Center of Educational Big Data, Central China Normal University, Wuhan, China\\
   $^2$Faculty of Artificial Intelligence in Education, Central China Normal University, Wuhan, China\\
   $^3$School of Software Technology, Zhejiang University, Hangzhou, China\\
   \texttt{\{liusy027, zkyang027, wanq8228, shenxiaoxuan, sunjw\}@ccnu.edu.cn}\\
   \texttt{fjt2018@mails.ccnu.edu.cn}, 
   \texttt{yaweiluo@zju.edu.cn}
}

\maketitle
\vspace{5cm}  % 添加 1 厘米的垂直空间

\def\customfootnotetext#1#2{{%
         \let\thefootnote\relax
         \footnotetext[#1]{#2}}}

% \customfootnotetext{1}{\textsuperscript{}Work in progress}
\customfootnotetext{1}{\textsuperscript{*}Qian Wan, Xiaoxuan Shen, and Jianwen Sun are the corresponding authors.}

\begin{abstract}
The automatic generation of high-quality mathematical problems is practically valuable in many educational scenarios. Large multimodal model provides a novel technical approach for the mathematical problem generation because of its wide success in cross-modal data scenarios. However, the traditional method of separating problem solving from problem generation and the mainstream fine-tuning framework of monotonous data structure with homogeneous training objectives limit the application of large multimodal model in mathematical problem generation. Addressing these challenges, this paper proposes COMET, a ``Cone of Experience'' enhanced large multimodal model for mathematical problem generation. Firstly, from the perspective of mutual ability promotion and application logic, we unify stem generation and problem solving into mathematical problem generation. Secondly, a three-stage fine-turning framework guided by the ``Cone of Experience'' is proposed. The framework divides the fine-tuning data into symbolic experience, iconic experience, and direct experience to draw parallels with experiences in the career growth of teachers. Several fine-grained data construction and injection methods are designed in this framework. Finally, we construct a Chinese multimodal mathematical problem dataset to fill the vacancy of Chinese multimodal data in this field. Combined with objective and subjective indicators, experiments on multiple datasets fully verify the effectiveness of the proposed framework and model.
\end{abstract}

\textbf{keywords:} Mathematical Problem Generation, Cone Methodology, Large Multimodal Model, Educational Large Model, Smart Education

%%%%%%%%%%%%%%%%%%%%%%%%%%%%%%%%%%%%%%%%%%%%%%%%%%%%%%%
%%% The main text. 正文部分
%%%%%%%%%%%%%%%%%%%%%%%%%%%%%%%%%%%%%%%%%%%%%%%%%%%%%%%
\section{Introduction}
As a vital driving force leading the revolution of technological and industrial development, generative artificial intelligence (GenAI) is restructuring various industries. For education, the impact of GenAI is unprecedented \cite{wang2023scientific}. The Large Language Model (LLM), as one of the most representative technologies of GenAI, displays excellent capabilities in text generation and processing \cite{achiam2023gpt,ouyang2022training,sun2023moss}. The Large Multimodal Model (LMM) further expands the data boundaries of the LLM and has achieved widespread success in cross-modal tasks (including image captioning and visual question answering) \cite{llava_liu2024visual,li2023monkey,chen2023internvl}. The integration and application of LMM has become a key approach to promote the digital transformation of education, since most of the teaching resources and records in the educational scenario are multimodal data.

\begin{figure*}[h]
	\centering
	\includegraphics[width=\linewidth]{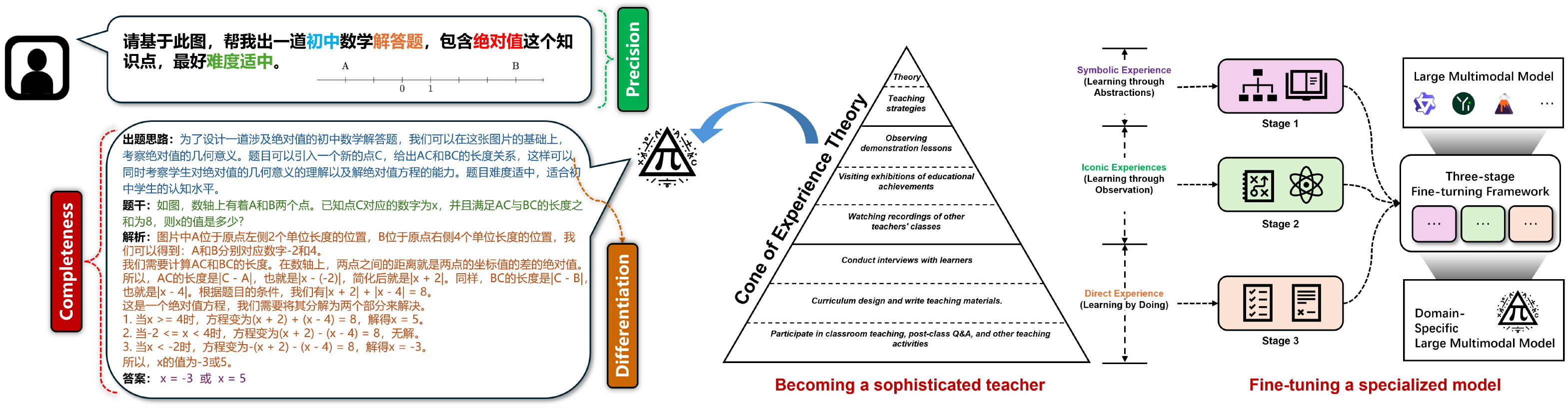}
	\caption{The diagram of mathematical problem generation and the ``Cone of Experience'' guided model fine-tuning.}
	\label{fig1}
\end{figure*}

In recent years, many researchers have been exploring the possibilities of combining LMM with education, such as assisted writing\cite{liu2024investigating} and emotional support\cite{lissak2024colorful}. However, there is still a lack of relevant research in the generation of educational resources, especially in the field of mathematical problem generation. The shortage of high-quality educational resources is one of the main contradictions in the digitization of education. As shown in Figure~\ref{fig1}, a high-quality mathematical problem needs to be carefully designed by domain experts and meet multiple requirements. First of all, completeness. During the teaching process, the mathematical problem is for teachers, students, and parents concurrently. Therefore it should contain four parts: mind of design, stem, mind of solution, and answer, all with fluent language and correct logic. Secondly,  precision. The mathematical problem should accurately reflect the objectives of the curriculum, be highly related to given knowledge points, and provide the function of exercises and tests. Lastly, differentiation. For certain key knowledge points under investigation, the mathematical problem should differentiate in theme, problem type, difficulty level, etc., to better serve complex and diverse learning needs.

In summary, constructing high-quality mathematical problems requires the ability to generate both stems and solutions to form a complete closed loop. Traditionally, the studies of mathematical problem generation are divided into two independent subfields, namely stem generation\cite{polozov2015personalized} (some works simply record as problem generation) and problem solving\cite{kushman2014learning}. 
These studies mostly design rules or deep neural networks to achieve reasoning, but are ineffective due to limitations in feature engineering and model capabilities, and the research paradigm that separates stem generation and problem solving does not meet the application requirement in educational scenarios. 
LLM, which provides a novel approach for mathematical problem generation, can not only generate coherent and logical replies against cross-modality data, but also respond to diverse demands because of its ability to in-context learning and instruction following. 
However, there are still challenges when directly applying the existing LMM to mathematical problem generation. Firstly, current work mostly focuses only on enhancing one aspect of the abilities of LLM in stem generation or problem solving, with little research proposing methods to simultaneously enhance both aspects of the model on the scale of multimodality. Secondly, general LMMs have learned abundant general concepts from a massive amount of pre-training data, but lack specialized knowledge needed for mathematical problem generation. Thirdly, implementing domain fine-tuning based on general LMM is currently the basic paradigm to execute domain task transfer. Previous data structure, as well as construction methods, are simple, and the training objectives are not diverse enough, thus it's hard to fully adapt to the application requirements of the target domain.

To address the above issues, this paper proposes a ``Cone of Experience'' enhanced large multimodal model for mathematical problem generation (COMET). Firstly, stem generation and problem solving are unified into mathematical problem generation tasks. Intuitively, the professional knowledge and practical experience required for stem generation and problem solving share commonalities. Integrating the two abilities into a single model can benefit the promotion of each other, and is more practically logical in educational scenarios. 
Secondly, inspired by the ``Cone of Experience'' theory proposed by American educator Edgar Dale \cite{dale1947audio}, we propose a three-stage fine-turning framework. The ``Cone of Experience'' divides human learning experience into three layers: symbolic experience, iconic experience, and direct experience. The experiences of different layers are interconnected and only by fully integrating all three layers can high-quality learning be achieved. From the perspective of Data-centric AI (DCAI) \cite{zha2023data}, we believe that the depth and breadth of transfer training are key to domain transfer. Accordingly, for the specific task of mathematical problem generation, we design multiple fine-grained data production methods for the three types of experiences, establish multi-level experience data injection methods, and form a complete fine-turning framework. Finally, a Chinese multimodal mathematical problem dataset (CMM12K) is formulated, filling the gap in Chinese multimodal corpus in this field. The effectiveness of the framework and model is comprehensively validated with both objective and subjective indicators on multiple datasets.

The main contributions of this paper can be summarized as follows:
\begin{itemize}
	\item From the perspective of DCAI, we propose COMET, a ``Cone of Experience'' enhanced large multimodal model for mathematical problem generation. To the best of our knowledge, this is the first work to systematically enhance mathematical problem generation on a single LMM.
	
	\item The formal definition of a three-stage fine-tuning framework based on the ``Cone of Experience'' is provided, together with the data flow production methods for symbolic experience, iconic experience, and direct experience. The corresponding knowledge infusion methods are empirically demonstrated.
	
	\item A Chinese multimodal mathematical problem dataset (CMM12K) is built, which includes 4 types of problems and $12,000$ samples. This work fills the gap in the field of Chinese multimodal corpus and provides a high-quality benchmark for subsequent research.
% 下面这个private感觉怪怪的	 private and public
	\item A large number of experiments have been carried out on multiple datasets, and the advancement and effectiveness of the proposed framework and model have been verified through qualitative and quantitative analysis.
\end{itemize}

\section{Related Work}
As mentioned above, the complete closed-loop of mathematical problem generation involves two dimensions of stem generation (previous works simply record as problem generation) and problem solving. This section introduces related work from the perspective of technological development.

Early studies focus on the design of generation rules and reason templates through summarizing the characteristics and patterns of mathematical problems\cite{singh2012automatically}. These works accomplish stem generation or mathematical reasoning by combining the concepts, formulas, and theorems \cite{polozov2015personalized}. Nandhini et al. \cite{nandhini2011math} proposed two stem generation methods based on templates and context-free grammar, generating stem with more diversified structures and semantics. Moura et al. \cite{de2015lean} built a knowledge base with a large number of mathematical theorems embedded, providing interactive theorem proving based on rule reasoning. These methods have a certain degree of controllability and a high accuracy rate, but lack problem adaptability and creativity.

With the development of machine learning, methods such as decision trees and support vector machine have been widely used to address mathematical stem generation or reasoning, recognizing the structure and patterns of problems by models which are trained based on large amounts of labeled data. Heilman et al. \cite{heilman2011automatic} used a syntactic parser to convert input text into tree representations and designed templates to achieve automatic transformation of problem forms. Roy et al. \cite{roy2015reasoning,roy2016solving} mapped unstructured text to a more easily reasoned representation space to eliminate text ambiguity, which can reason multi-step arithmetic problems. Deep learning further provides more powerful models, including seq2seq model, attention mechanism, graph network, etc., providing new approaches for the generation and reasoning of a wide range of mathematical problems such as arithmetic, algebra, and geometry \cite{wu2022automatic,liu2019tree,cao2021bottom,wang2018translating,chen2021geoqa}. Zhou et al. \cite{zhou2019towards} first proposed a seq2seq model based on the attention mechanism, which generated the stem of applied problems given equations and mathematical topics, significantly improving the quality and diversity of generation. Wang et al. \cite{wang2017deep} used recurrent neural network to transform math word problems into equations and based on similarity retrieval to improve reasoning performance. Group-ATT \cite{li2019modeling} applied multi-head attention to extract the global feature, numerical feature, and quantity pair feature of math word problems, achieving significant performance improvement on the Math23K dataset.

The mathematical problem generation has entered a new stage because of the sharply developed and applied LLM \cite{christ2024mathwell,drori2022neural,yue2023mammoth,zhou2023solving}, since by training on large amounts of corpus LLMs can understand complex language structures including mathematical problems. In terms of stem generation, Droria et al. \cite{drori2022neural} use OpenAI Codex (LLM of code data fine-tuning) to approach human-level in generating college-level mathematical stems. Zong et al. \cite{zong2023solving} based on the few-shot learning prompt GPT-3 to achieve the generation of related topics. In terms of problem solving, WizardMath \cite{luo2023wizardmath} proposed a reinforcement learning from the evol-instruct feedback method to construct more complex instruction datasets, enhancing the mathematical reasoning ability of LLaMA-2. MathGLM \cite{yang2023gpt} demonstrates proficient multi-digit arithmetic ability with only 2B parameters. MathPrompter \cite{imani2023mathprompter} increases the credibility of the output result by generating multiple algebraic expressions or Python functions to solve the same problem based on zero-shot chain of thought. ToRA \cite{gou2023tora} integrates computational libraries and symbolic solvers to solve complex mathematical problems.

\section{Methodology}
\begin{figure*}[h]
	\centering
	\includegraphics[width=\linewidth]{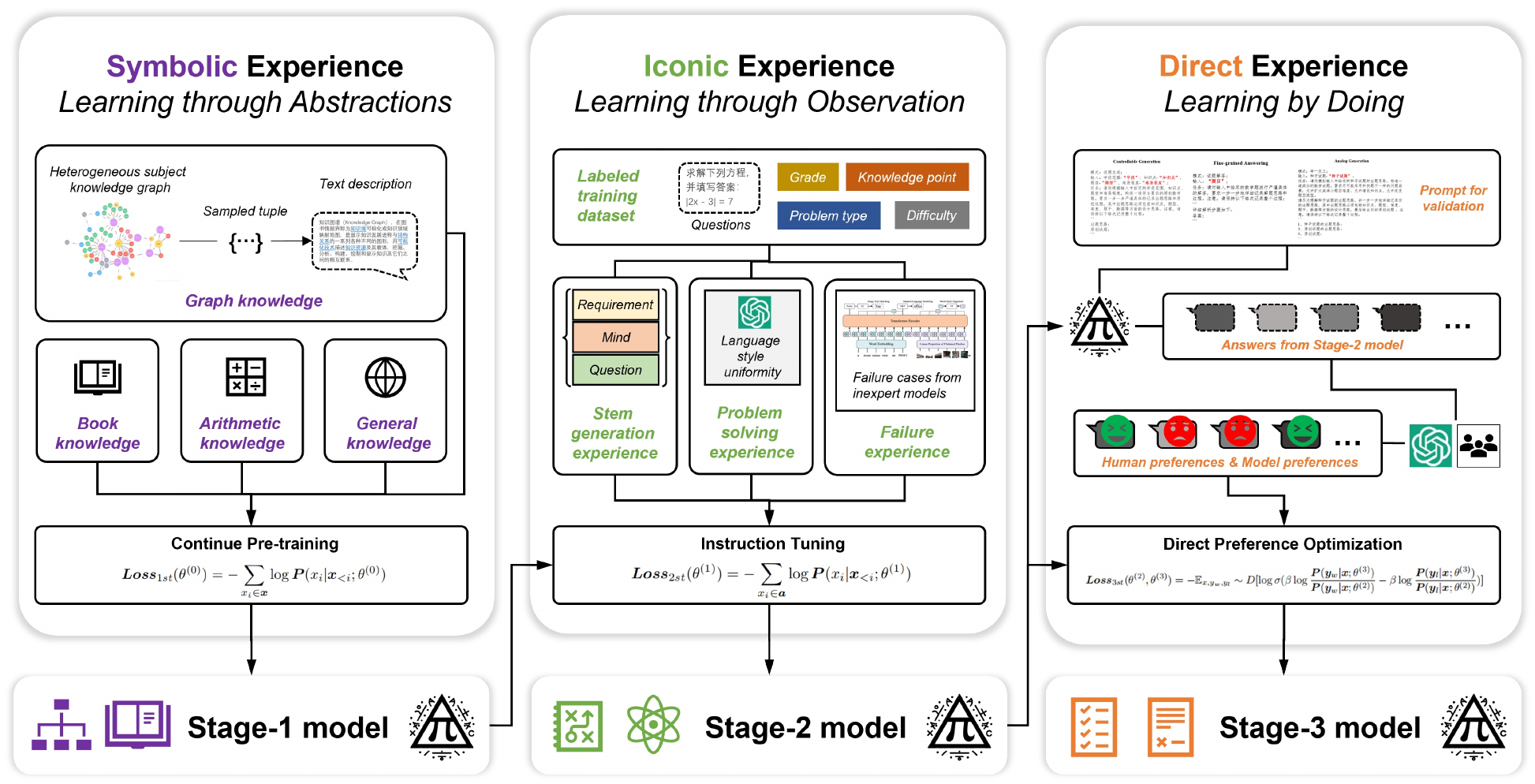}
	\caption{The diagram of the three-stage fine-tuning framework.}
	\label{fig2}
\end{figure*}

Figure~\ref{fig2} is a schematic diagram of the three-stage fine-tuning framework. The entire fine-tuning process is guided by the ``Cone of Experience'', injecting symbolic experience, iconic experience, and direct experience. This section first defines the global fine-tuning goals and notations, decomposing the application requirements of the target domain into three subtasks for reinforcement. Then, the three-stage fine-tuning framework is expanded according to the type of injected experience, elaborating on the definitions, construction methods, and training methods.

% \vspace{-1 \baselineskip} % 调整间距大小
\subsection{Problem Formulation}
To effectively apply LMM in teaching scenarios, this work mainly enhances three capabilities of LMM during the domain fine-tuning process: controllable generation (CG), analogy generation (AG), and fine-grained solving (FS) for mathematical problems. Both CG and AG reflect the ability of LMM to generate problems, the difference being that the former generates the mind of design and original problem according to given requirements (such as problem type, knowledge point, difficulty level, etc.), while the latter understands and transforms the seed problems (such as changing topic and type, expanding knowledge point or adjusting difficulty level). The FS reflects the problem-solving capacity of LMM, emphasizing the importance of producing detailed solution steps similar to textbook references.

For LMM, the instructions for the above three tasks can be formally defined as follows:

% \vspace{-0.5 \baselineskip} % 调整间距大小、
\begin{enumerate} % [itemsep=2pt,topsep=0pt,parsep=0pt]

    \item Given the problem type $\bm{t}$, knowledge point $\bm{c}$, difficulty-level $\bm{d}$ and grade level $\bm{g}$, the CG prompt is constructed as $\bm{q}_{c}=\bm{F}_{c}(\bm{t},\bm{c},\bm{d},\bm{g})$.
    
    \item Given the seed problem $\bm{s} \in S$, the AG prompt is constructed as $\bm{q}_{a}=\bm{F}_{a}(\bm{s})$.
    
    \item Suppose a math problem is $\bm{p}$, the prompt identifier of FS is $\bm{q}_{s}=\bm{F}_{s}(\bm{p})$.
\end{enumerate}
\vspace{-0.5 \baselineskip} % 调整间距大小

The $\bm{F}_{c}, \bm{F}_{a}, \bm{F}_{s}$ can be flexibly designed according to the scene, and the settings in this work can be seen in Section 3.4. Please note that in this paper, $\bm{x}$ represents a vector or a string, $x$ represents a scalar or a single character, $X$ represents a set, and $\bm{X}$ represents a function.

The task requirements are defined as $\bm{q}_{in} \in \{\bm{q}_{c},\bm{q}_{a},\bm{q}_{s}\}$. This work can be defined as performing three-stage fine-tuning based on the general LMM $\bm{F}_{lmm}^{(0)}$, combined with the ``Cone of Experience'' theory, to obtain an LMM $\bm{F}_{lmm}^{(3)}$ that meets the application requirements of mathematical problem generation in the teaching scene, so as to maximize the following conditional probability:

\begin{multline}
    \bm{P}_{lmm}(\bm{m}|\bm{q}_{in};\theta^{(3)})=\\
    \prod_{k=1}^{N_m}\bm{P}_{lmm}(w_k^m|\bm{q}_{in} \oplus \bm{w}_{<k}^m;\theta^{(3)}).    
\end{multline}
\vspace{-0.3cm}
\begin{multline}
    \bm{P}_{lmm}(\bm{a}|\bm{q}_{in},\bm{m};\theta^{(3)})=\\
    \prod_{k=1}^{N_a}\bm{P}_{lmm}(w_k^a|\bm{q}_{in} \oplus \bm{m} \oplus \bm{w}_{<k}^a;\theta^{(3)}).
\end{multline}

where $\theta^{(3)}$ is the parameters of LMM $\bm{F}_{lmm}^{(3)}$, $\oplus$ represents the string concatenation operation. $\bm{m}=\{w_1^m,w_2^m,\dots,w_{N_m}^m\}$ represents the mind of design or problem-solving steps generated by LMM, and $\bm{a}=\{w_1^a,w_2^a,\dots,w_{N_a}^a\}$ represents the original problem or final answer generated by the LMM.

\subsection{Symbolic Experience: Learning through Abstractions}
This paper defines symbolic experience as the background knowledge related to the target domain, or the prerequisite knowledge required to carry out the target task. Symbolic experience does not directly help the model solve specific tasks, but it provides strong support. For mathematical problem generation, we summarize symbolic experience into four types for production: book knowledge, graph knowledge, arithmetic knowledge, and general knowledge.

The data sources of \textbf{book knowledge} include textbooks, lecture notes, teachers’ books, pedagogy, and psychology books, aiming to build teaching concepts and supplement subject knowledge. Through methods such as web crawling, OCR, and manual annotation, we complete data collection and pre-processing (de-duplication, noise reduction, etc.) via both online and offline channels. The number of book knowledge tokens sorted out in this work is approximately 140M.

We construct a large heterogeneous subject knowledge graph, where the node types include grade, knowledge points, concept descriptions, and example problems. This graph encompasses $1,225$ knowledge points and related concepts from elementary to junior high school, providing approximately $18,000$ example problems. To train LMM using structured data, we design a graph sampling method based on random walk to extract diversified and differentiated disciplinary information. Then GPT4(V) is used to transform the edge information into a concatenated text, thereby generating \textbf{graph knowledge} for symbolic experiences. Specifically, the heterogeneous subject knowledge graph is represented as $G=<\{N_c, N_g, N_d, N_p\}, E>$, where $N_c, N_g, N_d, N_p$ represent the node sets of knowledge points, grade, concept descriptions, and related example problems. $E$ is the set of edges between all nodes. The generation process of graph knowledge can be seen in Algorithm~\ref{kgalg}, which generates two types of training samples: a whole link learning sample (Sample\_1) is formed as a four-tuple \textit{\{grade, knowledge point, concept description, example problem\}}, and a relationship learning sample (Sample\_2) formed by the concatenation of multiple adjacent knowledge points, totaling 220M tokens.

\begin{algorithm}[h]
	\centering
	\caption{Graph Knowledge Generation}
	\label{kgalg}
	% \begin{multicols}{2}
		\renewcommand{\algorithmicrequire}{\textbf{Input:}}%更改输入名称
		\renewcommand{\algorithmicensure}{\textbf{Output:}}%更改输出名称
		\footnotesize
		\begin{algorithmic}[1]
			\REQUIRE $G=<\{N_c, N_g, N_d, N_p\},E>$, \\$D_1=\emptyset, D_2=\emptyset$
			\ENSURE Sample\_1, Sample\_2%, Sample\_3
			\STATE $n_i=\bm{random}(N_c)$
			\STATE $D_1=D_2=\{n_c\}$
			\STATE $n_j=\bm{random}(N_g)$
			\IF{$e_{ij}$ is not None}
			\STATE $D_1=D_1\cup \{n_j\}$
			\ENDIF
			\STATE $n_j=\bm{random}(N_d)$
			\IF{$e_{ij}$ is not None}
			\STATE $D_1=D_1\cup \{n_j\}$
			\ENDIF
			\FOR{$k=1;k<3;k++$}
			\STATE $n_j=\bm{random}(N_d)$
			\IF{$e_{ij}$ is not None}
			\STATE $D_2=D_2\cup \{n_j\}$
			\ENDIF
			\ENDFOR
			\FOR{$k=1;k<5;k++$}
			\STATE $n_j=\bm{random}(N_c-\{n_i\})$
			\IF{$e_{ij}$ is not None}
			\STATE $D_2=D_2\cup \{n_j\}$
			\ENDIF
			\ENDFOR
			\STATE Sample\_1 = $\bm{GPT4V}(D_1)$
			\STATE Sample\_2 = $\bm{GPT4V}(D_2)$
		\end{algorithmic}
	% \end{multicols}
\end{algorithm}

The function of \textbf{arithmetic knowledge} is to compensate for the shortcomings of LMM in arithmetic, to reduce the probability of numerical errors occurring in the mathematical reasoning process. It is an equation consisting of pure numbers and mathematical operators. We directly use the arithmetic dataset proposed by Yang et al. \cite{yang2023gpt}. This dataset is carefully designed, containing not only operations such as addition, subtraction, multiplication, division, and exponentiation, but also various numerical formats such as integers, decimals, percentages, fractions, and negative numbers. In this work, approximately 200M tokens are extracted as fine-tuning data for arithmetic knowledge.

We extracted approximately 220M tokens of generic data (including plain text, single-turn, and multi-turn Q\&A) from open-source corpora, such as Wikipedia, SkyPile-150B\cite{wei2023skywork}, MOSS\cite{sun2023moss} and BELLE\cite{ji2023exploring}, as the \textbf{general knowledge} in symbolic experience. The main role of this knowledge is to slow down the forgetting phenomenon caused by continued pre-training.

This stage processes all the data associated with symbolic experience as pre-training form and infuses it into the LMM for learning, i.e., no masking of data content is undertaken. The backpropagation of model training computes loss from the first token of the input. Assuming the input sample is $\bm{x}$, the loss function at this stage is as follows:
\begin{equation}
	\bm{Loss}_{1st}(\theta^{(0)})=-\sum_{x_i \in \bm{x}}\log\bm{P}(x_i|\bm{x}_{<i};\theta^{(0)}).
\end{equation}

\subsection{Iconic Experience: Learning through Observation}
The iconic experience is defined as the data generated by the subject in the process of performing the target task, which includes not only human experts proficient in the target task but also the large model. Injecting the iconic experience aims to allow LMM to learn mathematical problem generation from humans and improve upon the failed reasoning data produced by other LMMs. This paper summarizes the iconic experience into three types of production: the experience of stem generation, problem solving, and failure.

To construct \textbf{stem generation experience}, we first collect exercises and test items covering all grades from elementary to junior high school. Next, based on manual annotation methods, we extracted key information from math problems in several dimensions including educational grade, problem type, knowledge points, and difficulty, and deduced problem requirements in reverse. Finally, we constructed a query-problem pair, with manual writing examples of mind of design, and used GPT4(V) for bulk supplementation of question making ideas in a few-shot manner. The final data form is \textit{\{problem requirement, mind of design, original problem\}}.

To construct \textbf{problem solving experience}, we hire normal school students to write analyses and answers for the collected mathematical problems. However, due to differences in cognitive levels and writing styles between individuals, it is difficult to align the granularity of the analyses. To generate fine-grained analyses, we use GPT4(V) to generate high-quality analyses with consistent writing styles based on manually parsed data. Three generation methods are proposed:

\vspace{-0.45 \baselineskip} % 调整间距大小
\begin{enumerate} % [itemsep=2pt,topsep=0pt,parsep=0pt]
    \setlength{\itemsep}{1pt}
    \setlength{\parskip}{0pt}
    \setlength{\parsep}{0pt}
    
    \item The task requires GPT4(V) to directly solve the problem: $\{\bm{q}\}\rightarrowtail\{\bm{s}\}$.
    \item The task requires GPT4(V) to fill in the middle process when both the problem and answer are given: $\{\bm{q},\bm{a}\}\rightarrowtail\{\bm{s}\}$.
    \item When the complete problem, analyses, and answer are given, GPT4(V) is required to rewrite the analyses: $\{\bm{q},\bm{s},\bm{a}\}\rightarrowtail\{\bm{s'}\}$.
\end{enumerate}
\vspace{-0.45 \baselineskip} % 调整间距大小

We chose the second method as the data production method for this stage due to its stability. The final data form is \textit{\{mathematical problem, mind of solution, final answer\}}.

\textbf{Failure experience} is mainly generated by LMMs that have not been domain-adapted. First, a collaborative environment consisting only of LMMs is built, among which GPT4(V) plays the role of the discriminator, and multiple LMMs (such as Qwen-VL-Chat, Yi-VL-6/34B, etc.) play the role of generators. Secondly, two generators are randomly assigned to complete the task of mathematical problem generation, and then the discriminator guides and evaluates the degree of completion. Finally, the summarized procedural data forms a sample in the format \textit{\{task instruction, collaboration information, guidance feedback\}}.

In this stage, the data pertaining to the iconic experience is learned by the LMM in the form of instruction Tuning. All data is arranged in a query-response pair, and a masking process is applied to the query part. The backward propagation of model training only starts calculating loss from the first token of the response. Suppose the query-response pairs are defined as $\bm{q}:\bm{a}$, the model input sequence is $\bm{x}=\{\bm{q}\oplus\bm{a}\}$. The loss function at this stage is as follows:
\begin{equation}
	\bm{Loss}_{2st}(\theta^{(1)})=-\sum_{x_i \in \bm{a}}\log\bm{P}(x_i|\bm{x}_{<i};\theta^{(1)}).
\end{equation}

\subsection{Direct Experience: Learning by Doing}
The direct experience is defined as the procedural data generated when the fine-tuned object carries out the target task with results feedback. Such experience aims to correct the inference preference of the LMM with higher-order domain values, allowing it to embodied evolve during the practice.

\begin{figure*}[h]
	\centering
	\includegraphics[width=0.9\textwidth]{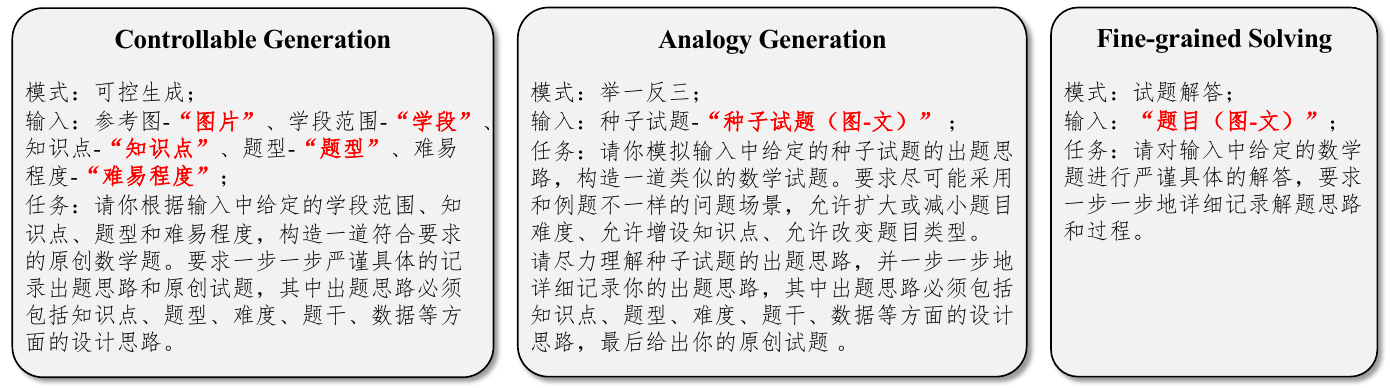}
	\caption{Prompt of the three tasks.}
	\label{prompt}
\end{figure*}

Firstly, we design a set of task instructions for three subtasks (CG, AG, and FS). For CG, the focus of the prompt design is to highlight the controllable elements in the generation process. This paper mainly considers four controllable factors in the problem generation process: grade, problem type, knowledge point, and difficulty level, and requires giving out the mind of design. For AG, the prompt design focuses on asking the model to first understand the seed problem to initially judge the important elements such as problem type and knowledge points, and then give the mind of design and rewrite the problem in the form of chain of thought. For FS, the core concept of the prompt design is to clearly require the model to generate a detailed analysis process rather than just outputting an answer. All prompts designed in this work for the three tasks are shown in Figure~\ref{prompt}.

Secondly, the LMM can produce multiple different responses to the same query due to the randomness of its reasoning. We order the preferences of multiple responses corresponding to the same instruction. This paper utilizes human preferences (manual annotation) and model preferences (GPT4(V) generation) during the preference ranking process. The final data form is \textit{\{task instruction, high preference response, low preference response\}}.

The fine-tuning stage uses the direct preference optimization \cite{dpo_rafailov2024direct} (DPO) algorithm to infuse direct experience into LMM, the loss function is as follows:

\begin{multline}
\bm{Loss}_{3st}(\theta^{(2)},\theta^{(3)}) = -\mathbb{E}_{x,y_w,y_l \sim D} [ \log \sigma (  \\
 \beta \log \frac{\bm{P}(\bm{y}_w|\bm{x};\theta^{(3)})}{\bm{P}(\bm{y}_w|\bm{x};\theta^{(2)})} - \beta \log \frac{\bm{P}(\bm{y}_l|\bm{x};\theta^{(3)})}{\bm{P}(\bm{y}_l|\bm{x};\theta^{(2)})} ) ].
\end{multline}

the $\theta^{(3)}$ uses $\theta^{(2)}$ as the initial solution, for the same input $\bm{x}$, $\bm{y}_w$ and $\bm{y}_l$ represent the preferred solution and the non-preferred solution.

\section{Experiments}
\subsection{Implementation}
We conduct the ``Cone of Experience'' enhanced three-stage fine-tuning based on the well-trained Qwen-VL-Chat\cite{qwen_bai2023qwen} provided by Alibaba Cloud. For all fine-tuning stages, Adam with different learning rates is used as the optimizer, and the gradient truncation threshold is set to 0.5. We incorporate a warmup ratio of 0.05 and employ the batch size of 64. To control overfitting, we apply a weight decay of 0.1. The Max token is uniformly set to $2,048$, and the data is spliced or truncated in the pre-processing stage to improve training efficiency or reduce information loss. In addition, we employ deepspeed with ZeRO-2 stage\cite{deepspeed_rajbhandari2020zero} to improve parallel efficiency for speed up training.

In the first and second stages, we use LoRA \cite{lora_hu2021lora} to perform parameter-efficient fine-tuning, then set rank, alpha and dropout to 16, 32 and 0.05. All linear layers (including the image encoder) of LMM except the head layer are designated to apply the LoRA adapter. Among them, the learning rate of the first stage is set to $2\times 10^{-5}$, and halved in the second stage. We use the DPO\cite{dpo_rafailov2024direct} algorithm to inject direct experience for learning reasoning preferences in the third stage. The learning rate is $5\times 10^{-5}$, the DPO smoothing value is 0.1. To ensure reproducibility, the random seed is set to 42 during the whole experiment. The three-stage fine-tuning is performed on 8 NVIDIA A800-80G. For one epoch, the three-stage fine-tuning takes about 200, 50, and 20 GPU hours. In the test stage, the inference parameters of LMM are uniformly set top\_k to 20, top\_p to 0.7, repetition\_penalty to 1, and temperature to 0.3.

\subsection{Dataset and Baseline}
% 参考文献不准有链接
The datasets used are shown in Table~\ref{dataset}. GSM8K\cite{gsm8K_dataset} is an English single-modal math word problem (MWP) dataset, containing $7,473$ training samples and $1,319$ test samples. It mainly tests the reasoning and arithmetic abilities of primary school math. TAL-SCQ5K-CN\footnote{https://github.com/math-eval/TAL-SCQ5K} is a Chinese single-modal multiple-choice problem (MCP) dataset for K12 math, including $3,000$ training samples and $2,000$ test samples.

This work builds a Chinese multi-modal math problem dataset CMM12K, which includes $6,000$ single-modal math problems and $6,000$ multi-modal math problems, covering most of the knowledge points in K12 math from primary school to junior high school. This dataset contains four types of problems: MCP, math fill-in-the-blank problem (MFP), MWP, and math proof problem (MPP). The training set of CCM12K is divided into $10,000$ samples, and the development set and test set are $1,000$ samples each.

Five open source LMMs are specified as baselines, covering different parameter levels, including Qwen-VL-Chat(7B)\cite{qwen_bai2023qwen}, Yi-VL-6B/34B\cite{yi_young2024yi}, LLaVA-1.6(7B)\cite{llava_liu2024visual}, CogVLM(17B)\cite{congvlm_wang2023cogvlm}. It should be noted that in the three-stage fine-tuning, all datasets are aligned with Chinese. The training sets of each dataset participate in the construction of iconic experience and the direct experience mainly depends on the development set. The test set is completely isolated from all fine-tuning stages.

\begin{table*}[h]
	\caption{Statistics of datasets.}
	\label{dataset}
	\footnotesize
	\centering
	% \begin{tabular*}{\textwidth}{llccccc}
	\begin{tabular}{llccccc}
          \toprule
		{\textbf{Dataset}} & \textbf{Language} & \textbf{Modal} &  \textbf{Type}  & \textbf{\#Train.} & \textbf{\#Dev.} &  \textbf{\#Test.} \\
		\hline
		GSM8K       & $\text{En} \rightarrow \text{Zh}$     & Single   & MWP  & $7,473$     & -    & $1,319$      \\
		TAL-SCQ5K-CN   & Zh       & Single     & MCP    & $3,000$     & -    & $2,000$     \\
		CMM12K        & Zh       & Multi    & MCP, MFP, MWP, MPP    & $10,000$     & $1,000$    & $1,000$   \\
		\bottomrule
	\end{tabular}
\end{table*}

\subsection{Metrics}
This study designs three types of evaluation criteria for different capabilities of LMM.

\textbf{Scoring mode based on GPT4(V)}. Multiple scoring dimensions are designed for \textbf{controllable generation} (CG), \textbf{analogy generation} (AG), and \textbf{fine-grained solving} (FS). The scoring dimensions of CG include language fluency (LF) (both mathematical terms and formulas), logical correctness (LC), content completeness (CC) (both ideas and stems), knowledge point relevance (KR), difficulty appropriateness (DA) and type adaptability (TA). The scoring dimensions of AG include language fluency (LF), logical correctness (LC), content completeness (CC), reasoning rationality(RR), and seed relevance (SR). The scoring dimensions of FS include language fluency (LF), logical correctness (LC), analytical completeness (AC), and answer accuracy (AA). The GPT4(V) is required to give a score and reason in the range of 1 to 10 according to the dimension.

\textbf{Arena mode based on GPT4(V)}. Considering the subjectivity of mathematical problem generation, GPT4(V) is introduced as a referee to comprehensively rule on different responses of the same query from aspects such as accuracy, fluency, and values. Specifically, we calculate a rating value for each LMM to represent the ability on a certain task. During the judging process by GPT4(V), the ELO rating algorithm\cite{Elo_rating, nips_elo_zheng2024judging} is used to update the rating value of the participating LMM. Assume the initial ELO rating value is $1,000$. In $M$ rounds of competition, two LMMs (called LMM-$x$ and LMM-$y$) are randomly selected to reply to the same query each time. According to the ruling results of GPT4(V), the rating value is calculated as follows:
\begin{equation}
	% \nonumber
	E_x = \frac{1}{1+10^{\left(R_x-R_y\right) / 400}},
\end{equation}

\begin{equation}
	R_i^{'} = R_i + K\cdot (\bm{PK}(i) - E_i).
	\label{ELO}
\end{equation}
where $R_x$ and $R_y$ respectively represent the rating values of 
LMM-$x$ and LMM-$y$ in the previous round, $E_x$ and $E_y=1-E_x$ represent the current expected rating value. $R_i^{'} \left(i\in \{x, y\}\right)$ represents the updated rating value of LMM-$i$, $\bm{PK}(i)\in\{0,1\}$ is a boolean function that identifies whether LMM-$i$ win in this round. $K$ represents the K-factor, which defaults to 4 and controls the change rate of rating.

\textbf{Objective evaluation indicators}. For problems with clear answers (this work refers to MCP, MFP, and MWP), the accuracy (ACC) is returned through matching to report the solving performance of LMM. For the MPP, BLEU-1/2/L\cite{bleu_papineni-etal-2002-bleu} and ROUGE-1/2/3/4\cite{rouge_lin2004rouge} are used to approximate testing.

\section{Result and Discussion}

\subsection{Performance of Controllable Generation}
\begin{figure*}[h]
	\centering
	\includegraphics[width=\linewidth]{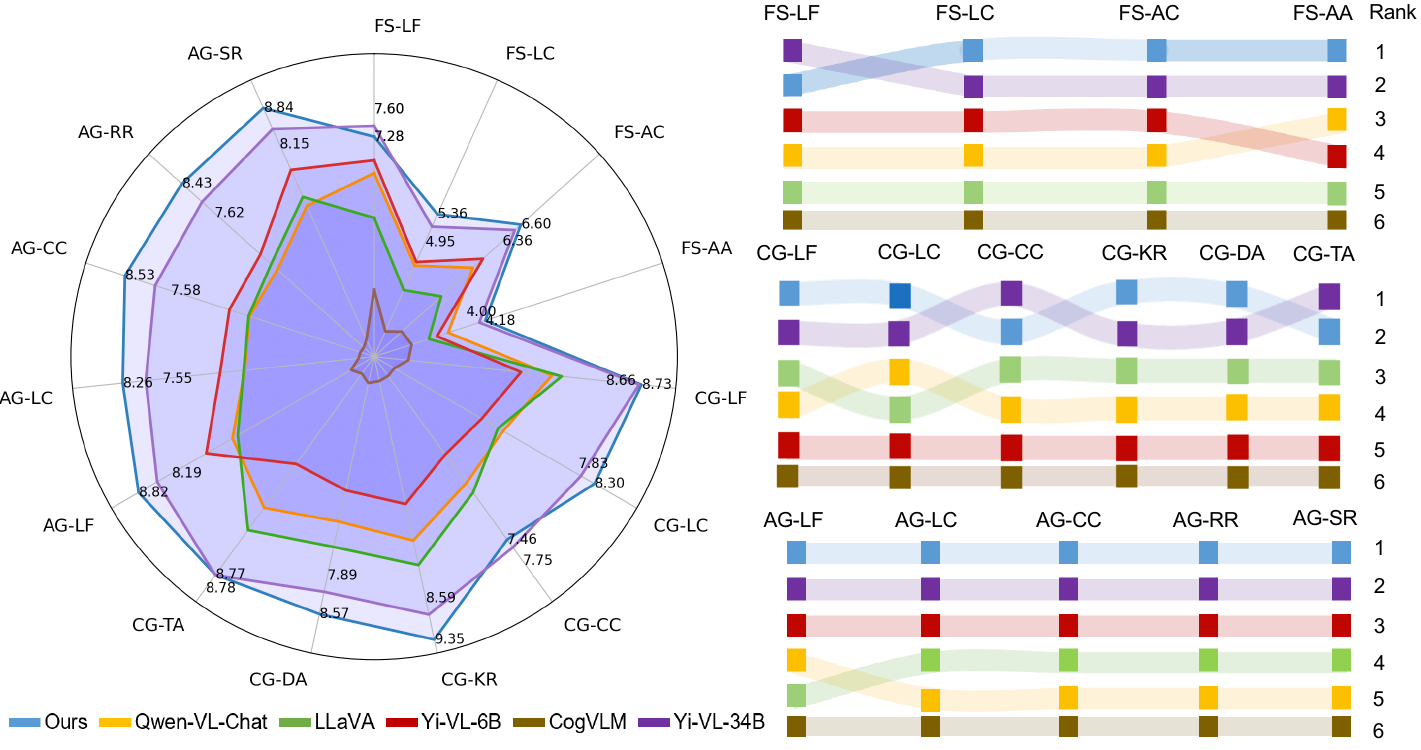}
	\put(-300,-5){\scriptsize(a)}
	\put(-100,-5){\scriptsize(b)} 
	
	\caption{Performance of our model and baselines on a broad range of problem-solving and generation fine-grained indicators. (a) The average score on 15 indicators in three tasks(CG, AG, and FS); (b) The rank of models' scores in each evaluation indicator.
    Here, each evaluation indicator is expressed as `task-dimension', and the number in the radar chart refers to the average score of models, we only present the scores of our model and Yi-VL-34B for visualization purposes. For instance, the label `CG-DA' depicted at the bottom of the radar chart denotes the difficulty appropriateness (DA) measure for the controllable generation (CG) task, where our model attains scores of 8.57 (rank 1), while Yi-VL-34B obtains 7.89 (rank 2).}
	\label{fig_score}
\end{figure*}

\textbf{Scoring mode}. We employed GPT4(V) to evaluate the responses of LMMs to CG tasks on the test set and subsequently reported the average scores across six dimensions. As illustrated in Figure~\ref{fig_score}(a), our model outperforms all other baselines in 4 of 6 dimensions(LF, LC, KR, and DA).

Figure~\ref{fig_score}(b) shows that our model performs comparably to Yi-VL-34B across the six dimensions of CG tasks. Models with comparable parameter levels, such as Qwen-VL-Chat (7B), LLaVA-1.6 (7B), and Yi-VL-6B, demonstrate slightly inferior performance, while the larger parameter-scale model, CogVLM with 17B parameters, exhibits relatively lower performance.

Although our model slightly lags behind Yi-VL-34B in terms of content completeness and problem type adaptability, it's important to note that our parameter count is approximately five times smaller than that of Yi-VL-34B. Therefore, this discrepancy may arise from limitations imposed by the scale of parameters, which could hinder the comprehension and processing of contextual information.

Through a three-stage fine-tuning process guided by the ``Cone of Experience'', our model can effectively focus on generating responses that prioritize controllable factors such as language fluency, logical correctness, knowledge point relevance, and difficulty appropriateness.

% FIG：雷达图

\textbf{Arena mode}. We conducted approximately $3,600$ competitions, where two LMMs were randomly selected for an anonymous battle, and the winner was determined by GPT4(V). As shown in Figure~\ref{fig_elo}(a), upon completing all competitions, our model's ELO rating significantly surpassed the baselines with a median value of $1,185$.

The heatmap illustrates the win rates of battles between various LMMs, revealing that our model exhibits absolute superiority when battling against Qwen-VL-Chat, Yi-VL-6B, and LLaVA-1.6. Moreover, even when pitted against Yi-VL-34B, our model maintains a slight edge with a 2\% advantage.

This outcome suggests that while larger parameter sizes may enhance text comprehension and generation capabilities, our model, after undergoing multi-modal fine-tuning guided by the ``Cone of Experience'', exceeds the performance of a 34B-parameter LMM in CG task without altering its parameter scale.

% ======================================
\subsection{Performance of Analogy Generation} % 举一反三
\textbf{Scoring mode}. In the AG task, our model demonstrates robust capabilities and absolute superiority. As depicted in Figure~\ref{fig_score}(a) and (b), firstly, our model comprehensively outperforms all baselines in various evaluation dimensions of AG, particularly surpassing the strong baseline Yi-VL-34B with a parameter size five times larger than ours. Secondly, our backbone model Qwen-VL-Chat ranks relatively lower among the six models, around the 4th to 5th position, trailing behind models with similar parameter scales such as Yi-VL-6B and LLaVA-1.6. However, following the three-stage fine-tuning guided by the ``Cone of Experience'', its AG capabilities have significantly improved, with an average score increase of approximately 60\% across all dimensions.

\begin{figure*}[h]
	\centering
	\includegraphics[width=\linewidth]{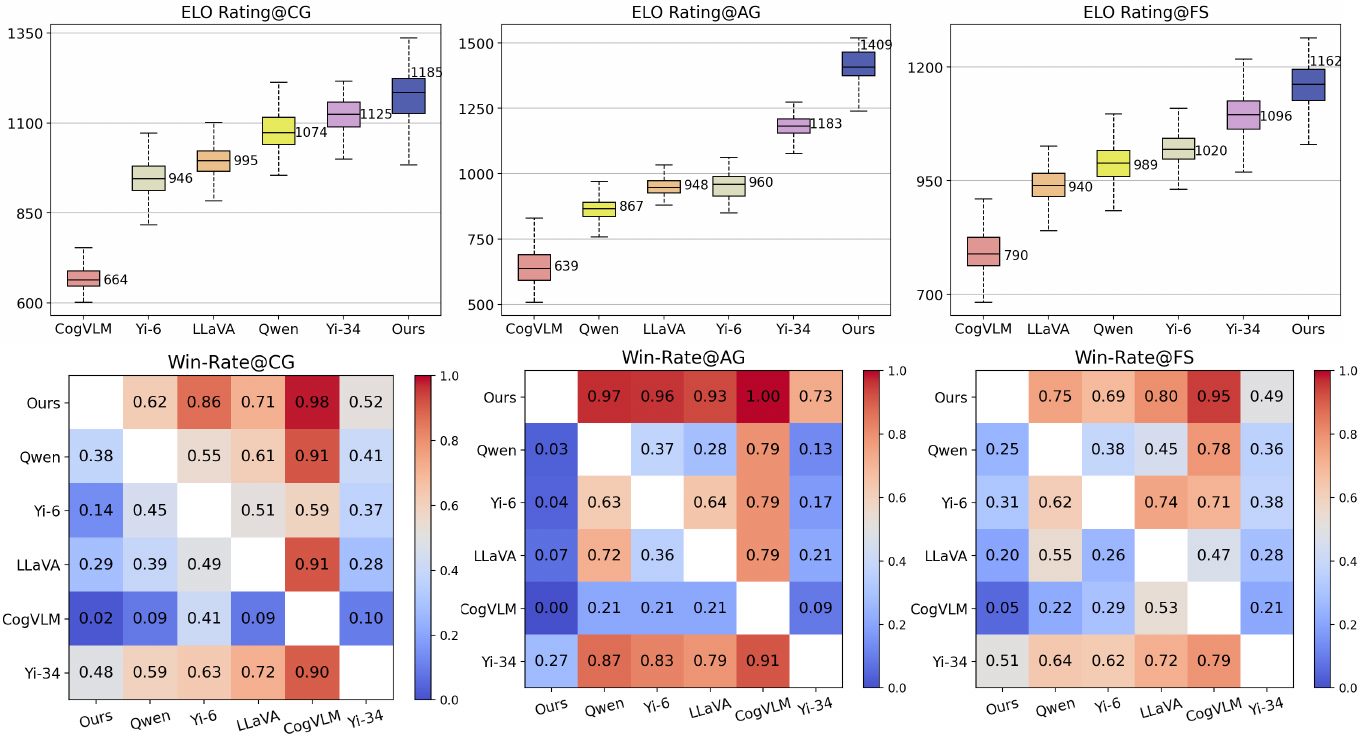}
	\put(-325,-5){\scriptsize(a)}
	\put(-195,-5){\scriptsize(b)} 
	\put(-65,-5) {\scriptsize(c)} 
	
	\caption{The statistics of ELO rating over $3,600$ rounds and the win rate between models. 
		The three subfigures, (a), (b), and (c), respectively represent tasks FS, CG, and AG. For each subfigure, the top section represents the ELO rating, here we sorted the models based on their ELO rating medians, while the bottom section represents the win rate.
        Abbreviations for Yi-VL-6B/34B, LLaVA1.6, and Qwen-VL-Chat, denoted as Yi-6/34, LLaVA, and Qwen respectively.
  }
	\label{fig_elo}
\end{figure*}

\textbf{Arena mode}. As illustrated in Figure~\ref{fig_elo}(b), after undergoing $3,600$ random competition rounds, our model's ELO Rating in AG tasks significantly surpasses all baselines. From the win rate heatmap, it's evident that our model exhibits absolute dominance when facing models with equivalent or double the parameter scale, defeating Qwen-VL-Chat, Yi-VL-6B, CogVLM, and LLaVA-1.6 with win rates of $97\%, 96\%, 100\%, 93\%$ respectively. Even when confronted with baselines five times larger in parameter size, our model maintains a win rate of 73\%. AG tasks hold educational value in demand for problem generation. Although our backbone model may lack training in this specific skill, our model successfully acquires it through the three-stage fine-tuning process. This outcome further validates the superiority of the proposed framework and model in this study.

% Fig: ELO Rating变化图、热力图

% ======================================
\subsection{Performance of Fine-grained Sovling}% 解题
\textbf{Scoring mode}. We report the scores for 4 dimensions in the FS task of all LMMs on the test set. The average scores across various evaluation dimensions are presented in Figure~\ref{fig_score}(a), where our model achieved the state-of-the-art (SOTA) in 3 of 4 evaluation dimensions(LC, AC, and AA), falling slightly short only in language fluency (LF) compared to Yi-VL-34B. Figure~\ref{fig_score}(b) depicts the ranking across all evaluation dimensions, further validating the superiority of our model, which maintains an absolute lead in most dimensions with a relatively smaller parameter size (7B). Moreover, by comparing with the backbone model Qwen-VL-Chat, we can conclude that the three-stage fine-tuning guided by the ``Cone of Experience'' significantly enhances its FS capabilities.
% , more details refer to \ref{sec_case_study}.

\textbf{Arena mode}. As illustrated in Figure~\ref{fig_elo}(c), after $3,600$ competition rounds, our model triumphs over all baselines with the highest ELO Rating. The average ELO rating leads the backbone model Qwen-VL-Chat by approximately 20\%, and also maintains an advantage against Yi-VL-34B. However, analyzing from the perspective of win rates reflected in the heatmap, our model lags behind Yi-VL-34B with a slight disadvantage, as it has a win rate of only 49\%.

Based on the scoring results of GPT4(V) and the arena outcomes, it can be concluded that, in the FS task, our model demonstrates significant performance improvement compared to the backbone model Qwen-VL-Chat, surpassing baselines of the same parameter scale comprehensively, and even shows certain advantages compared to baselines approximately five times its size.

% GSM, TAL, CMM12K所有客观题评价
\begin{table*}[h]
	\caption{Performances of each LMM on GSM8K, TAL-SCQ5K-CN, and CMM12K. Here Single/Multi indicates single modal/multi modal. All results are reported in terms of Acc (\%), \textbf{bold} indicates optimal performance while \underline{underline} indicates suboptimal performance.} 
	\label{tab_acc}
	\footnotesize
	\tabcolsep 8 pt %space between two columns. 用于调整列间距
	\centering
	\begin{tabular*}{\textwidth}{@{}cccccccccc@{}}
		\hline
		\multirow{3}{*}{\textbf{Model}} & \multirow{3}{*}{\textbf{GSM8K}} & \multirow{3}{*}{\textbf{TAL-SCQ5K-CN}} & \multicolumn{7}{c}{\textbf{CMM12K}}                                                                                                   \\ \cline{4-10} 
		&                        &                            & \multirow{2}{*}{Total} & \multicolumn{2}{c}{MCP}         & \multicolumn{2}{c}{MFP}         & \multicolumn{2}{c}{MWP}         \\ \cline{5-10} 
		&                        &                            &                        & Single              & Multi              & Single              & Multi              & Single         & Multi              \\ \hline
		Qwen-VL-Chat                & 18.04                  & \underline{16.20}                      & 24.12                  & \underline{27.33}          & 22.00          & 29.61          & 17.33          & 29.13          & 19.33          \\
		Yi-VL-6B               & 22.52                  & 6.55                       & 21.05                  & 14.00          & 18.67          & 32.19          & 18.67          & 28.16          & 14.67          \\
		LLaVA-1.6                  & 18.65                  & 10.05                      & 19.42                  & 17.33          & \underline{24.67}          & 27.04          & 11.33          & 22.82          & 13.33          \\
		CogVLM                 & 8.87                   & 5.35                       & 14.11                  & 4.67           & 22.67          & 21.89          & 7.33           & 17.48          & 10.67          \\
		Yi-VL-34B              & \textbf{38.66}                  & 7.60                       & \underline{24.54}                  & 16.67          & 16.67          & \textbf{38.62}    & \underline{19.33}          & \underline {33.98}    & \underline{22.00}          \\ \hline
		Ours               & \underline{28.89}            & \textbf{22.05}             & \textbf{33.84}         & \textbf{35.33}    & \textbf{34.00} & \underline {35.62}    & \textbf{29.33} & \textbf{35.44} & \textbf{33.33} \\ \hline
	\end{tabular*}
\end{table*}

\subsection{Result of Objective Evaluation}
\textbf{Performance of Objective Problems}. As shown in Table~\ref{tab_acc}, our model dramatically outperforms the backbone model Qwen-VL-Chat in terms of accuracy on GSM8K and TAL-SCQ5K-CN ($+10.62\%, +5.85\%$), achieving state-of-the-art (SOTA) on TAL-SCQ5K-CN. Although Yi-VL-34B leads on GSM8K, its parameter size, which is 5 times larger than ours, implies greater training cost and time. 

On CMM12K, our model's overall score of $33.84\%$ remarkably exceeds all baselines, with approximately an $8\%$ performance advantage over the second-place Yi-VL-34B. Specifically, we conducted statistics on two modalities and three problem types, totaling $2 \times 3=6$ categories. The results show that our model achieved SOTA in 5 of 6 categories, only slightly lagging behind Yi-VL-34B in single modal MFP by a small margin ($-3\%$). Compared with the baseline of the same parameter size, our model leads in all types of problems. For CogVLM, which has twice the parameter size of ours, our model maintains a lead of more than $15\%$ in all tasks. In summary, our model achieves relatively excellent problem-solving performance in all types of problems with a smaller parameter size (approximately $20\%$ of Yi-VL-34B).

% 证明题

\begin{table*}[h]
    \caption{
        Performances of each LMM on CMM12K with MPP, \textbf{bold} indicates optimal performance (higher is better for all metrics).
    }
    \label{zmt_tab}
    \centering
    \footnotesize
    \begin{tabular}{cccccccc}
        \toprule
        \textbf{Category} & \textbf{Metric} & \textbf{Qwen-VL-Chat} & \textbf{Yi-VL-6B} & \textbf{LLaVA} & \textbf{CogVLM} & \textbf{Yi-VL-34B} & \textbf{Ours} \\
        \midrule
        \multirow{7}{*}{Total} 
        & ROUGE@1 & 0.38 & 0.32 & 0.35 & 0.20 & 0.38 & \textbf{0.80} \\
        & ROUGE@2 & 0.25 & 0.20 & 0.21 & 0.05 & 0.20 & \textbf{0.67} \\
        & ROUGE@L & 0.32 & 0.27 & 0.23 & 0.11 & 0.27 & \textbf{0.69} \\
        & BLEU@1 & 0.28 & 0.32 & 0.27 & 0.18 & 0.34 & \textbf{0.71} \\
        & BLEU@2 & 0.17 & 0.15 & 0.11 & 0.05 & 0.16 & \textbf{0.62} \\
        & BLEU@3 & 0.12 & 0.10 & 0.07 & 0.02 & 0.11 & \textbf{0.54} \\
        & BLEU@4 & 0.08 & 0.07 & 0.05 & 0.01 & 0.07 & \textbf{0.47} \\
        \midrule
        \multirow{7}{*}{Single-Modal} 
        & ROUGE@1 & 0.37 & 0.32 & 0.37 & 0.21 & 0.37 & \textbf{0.81} \\
        & ROUGE@2 & 0.24 & 0.18 & 0.21 & 0.06 & 0.20 & \textbf{0.68} \\
        & ROUGE@L & 0.30 & 0.26 & 0.22 & 0.12 & 0.25 & \textbf{0.67} \\
        & BLEU@1 & 0.24 & 0.29 & 0.25 & 0.18 & 0.30 & \textbf{0.69} \\
        & BLEU@2 & 0.14 & 0.14 & 0.10 & 0.05 & 0.14 & \textbf{0.60} \\
        & BLEU@3 & 0.09 & 0.09 & 0.06 & 0.02 & 0.09 & \textbf{0.52} \\
        & BLEU@4 & 0.07 & 0.06 & 0.04 & 0.01 & 0.06 & \textbf{0.46} \\
        \midrule
        \multirow{7}{*}{Multi-Modal} 
        & ROUGE@1 & 0.39 & 0.33 & 0.33 & 0.18 & 0.40 & \textbf{0.80} \\
        & ROUGE@2 & 0.26 & 0.20 & 0.20 & 0.04 & 0.23 & \textbf{0.66} \\
        & ROUGE@L & 0.34 & 0.28 & 0.25 & 0.10 & 0.31 & \textbf{0.67} \\
        & BLEU@1 & 0.31 & 0.33 & 0.29 & 0.18 & 0.38 & \textbf{0.68} \\
        & BLEU@2 & 0.18 & 0.15 & 0.13 & 0.05 & 0.19 & \textbf{0.59} \\
        & BLEU@3 & 0.13 & 0.10 & 0.08 & 0.02 & 0.13 & \textbf{0.51} \\
        & BLEU@4 & 0.09 & 0.07 & 0.05 & 0.01 & 0.09 & \textbf{0.43} \\
        \bottomrule
    \end{tabular}
\end{table*}

\textbf{Performance of MPP}. In regards to the MPP in the CMM12K dataset, we draw from the evaluation logic of machine translation, comparing each LMM's response with the standard answer and calculating BLEU and ROUGE scores. The standard answer here is derived from manual annotation, encompassing not only the geometric elements and their relationships but also the complete proof process. By calculating BLEU and ROUGE, we can approximately determine whether the output of the LMM is in accordance with mathematical grammar and proof logic. Table~\ref{zmt_tab} displays the response quality of our model and the baseline model on single-modal and multi-modal proof problems. The results indicate that our model is superior to all baselines as far as the response quality of the proof problems is concerned.

%%%%%%%%%%%%%%%%%%%%%%%%%%%%%%%%%%%%%%%%%%%%%%%%%%%%%%%%%%%%%%%%%%%%%%%%
\subsection{Ablation Study}

% xiao-rong-image
\begin{figure*}[]
	\centering
	\includegraphics[width=\linewidth]{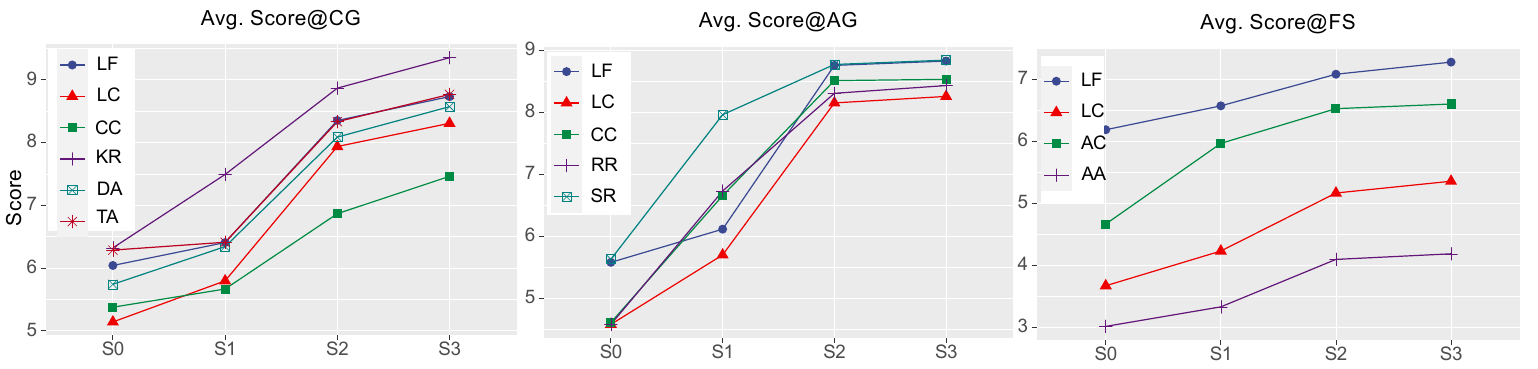}
	\put(-325,-5){\scriptsize(a)}
	\put(-185,-5){\scriptsize(b)} 
	\put(-65,-5) {\scriptsize(c)} 
	\caption{The average scores across various dimensions for our model at each stage of training.}
	\label{fig_ablation}
\end{figure*}

To validate the effectiveness of the three-stage fine-tuning framework proposed in this paper, we designed the following ablation experiments. We refer to the model after the $i^{th}$ stage of fine-tuning as Si, where $i\in \{1,2,3\}$. We preserved all checkpoints of the fine-tuning stages: S0 - the original backbone model Qwen-VL-Chat without any fine-tuning. S1 - the model after the first stage of fine-tuning, injected with symbolic experience. S2 - based on S1, the model after the second stage of fine-tuning, infused with iconic experience. S3 - based on S2, the model after the third stage of fine-tuning, incorporating direct experience. 

We first obtained these four models’ responses on the test set regarding the CG, AG, and FS tasks, and scored them based on GPT4(V) in 15 fine-grained dimensions. Figure~\ref{fig_ablation} shows the score changes on 15 dimensions for the three types of tasks. The results indicate that with the deepening of the three-stage fine-tuning, the model's scores in all dimensions show an increasing trend. 

We calculated the absolute performance improvements at each fine-tuning stage and reported them in Table ~\ref{tab_delta}. The results show that on most capability dimensions, $\Delta_2 = \max\{\Delta_1, \Delta_2, \Delta_3\}$, meaning that the second stage contributed the most to the performance improvement in the three-stage fine-tuning framework based on the ``cone of experience''.
We also observe some exceptions, including the four dimensions FS-EC, AG-CC, AG-RR, and AG-SR, whose common feature is that the original backbone model performs poorly, and the first stage of training plays a key role in improving the performance on these dimensions. In all dimensions of each task, further improvements in model performance can be achieved through continuous injecting of direct experience in the third stage of fine-tuning.

\begin{table}[h]
    \centering
    \caption{Performance improvements($\Delta$) of each evaluation dimension at each stage. Here, $\Delta_i$ represents the improvement in average score relative to the $(i-1)^{th}$ stage after the $i^{th}$ stage training, the \textbf{bold} indicates $\max\{\Delta_1, \Delta_2, \Delta_3\}$.}
    \label{tab_delta}
    \footnotesize
    \begin{tabular}{ccccc}
        \toprule
        Task & dimension  & $\Delta_1$ & $\Delta_2$ & $\Delta_3$ \\
        \hline
        \multirow{4}{*}{\textbf{FS}} & \textbf{LF} & 0.385 & \textbf{0.510} & 0.195 \\
                                     & \textbf{LC} & 0.560 & \textbf{0.935} & 0.190 \\
                                     & \textbf{AC} & \textbf{1.300} & 0.560 & 0.075 \\
                                     & \textbf{AA} & 0.315 & \textbf{0.765} & 0.090 \\
        \hline
        \multirow{6}{*}{\textbf{CG}} & \textbf{LF} & 0.365 & \textbf{1.945} & 0.380 \\
                                     & \textbf{LC} & 0.655 & \textbf{2.140} & 0.370 \\
                                     & \textbf{CC} & 0.290 & \textbf{1.200} & 0.595 \\
                                     & \textbf{KR} & 1.170 & \textbf{1.375} & 0.485 \\
                                     & \textbf{DA} & 0.600 & \textbf{1.745} & 0.485 \\
                                     & \textbf{TA} & 0.125 & \textbf{1.920} & 0.435 \\
        \hline
        \multirow{5}{*}{\textbf{AG}} & \textbf{LF} & 0.535 & \textbf{2.640} & 0.070 \\
                                     & \textbf{LC} & 1.120 & \textbf{2.450} & 0.105 \\
                                     & \textbf{CC} & \textbf{2.045} & 1.855 & 0.020 \\
                                     & \textbf{RR} & \textbf{2.150} & 1.575 & 0.125 \\
                                     & \textbf{SR} & \textbf{2.330} & 0.810 & 0.070 \\
        \bottomrule
    \end{tabular}
\end{table}

\subsection{Case Study}\label{sec_case_study}

% case study image
\begin{figure*}[!t]
	\centering
	\includegraphics[width=\linewidth]{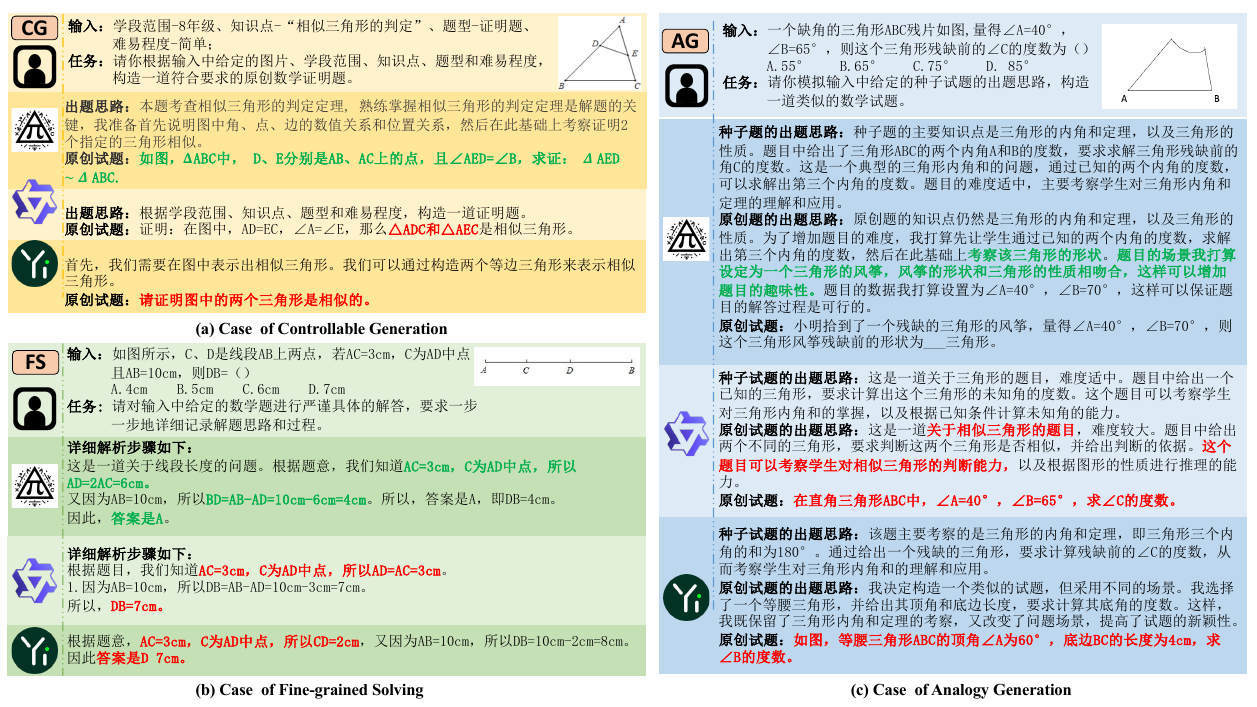}
	\caption{Case of the three tasks. The key correct parts of responses are highlighted in green and the incorrect ones in red.}
	\label{fig_case_study}
\end{figure*}

To demonstrate the strength of our model, we have selected some examples of CG, AG, and FS tasks respectively. Previous results, such as Figure~\ref{fig_score}, have already proven that Yi-VL-34B - a model with more than five times the number of parameters as ours, is a comparable competitor. Thus, for each task, we show the difference in response quality between our model, the backbone model Qwen-VL-Chat, and Yi-VL-34B when using the same prompt, the case details refer to Figure~\ref{fig_case_study}.

% ======== CG的例子 ======== 
% \vspace{-1 \baselineskip} % 调整间距大小
\paragraph{Case of CG} In the example shown in Figure~\ref{fig_case_study}(a), LMM was asked to generate one problem based on the given planar geometric picture. Our model accurately captured the geometric elements in the picture and expressed the test problem using the correct mathematical language. 
In contrast, Qwen-VL-Chat failed to comprehend the content of the given picture, erroneously providing the condition ($\triangle ADC \sim \triangle AEC$,  but both ADC and AEC are not triangles). For Yi-VL-34B, the problem it constructed was not based on the given image, hence not aligning with requirements.

% ======== FS的例子 ======== 
% \vspace{-1 \baselineskip} % 调整间距大小
\paragraph{Case of FS} Given this problem, Qwen-VL-Chat correctly understood the elements in the picture, but its erroneous reasoning steps (AD=AC=3cm is given, but actually AD=2AC=6cm) led to an incorrect final result. Yi-VL-34B made similar mistakes as Qwen-VL-Chat. However, our model first parsed the problem requirements, and then extracted the geometric elements of the given picture, and finally correctly reasoned step by step according to the problem to arrive at the correct answer (DB=4cm).

% ======== AG的例子 ======== 

\paragraph{Case of AG} We require LMMs to simulate the seed problem and construct a new problem. Each LMM first analyzes the ideas for the construction of the seed problem and then constructs a problem, and they are also asked to explain the thought process behind the constructed problem. 

Our model first understands the meaning of the problem, parses the content of the knowledge points tested (the sum of the interior angles of a triangle is 180$^\circ$), and tests a similar knowledge point (determine the shape of a triangle based on its angles) by modifying the problem scenario. 
This demonstrates that the symbolic experience, especially graph knowledge injected in the first stage helps the model find similar knowledge points based on the given knowledge point. 

For Qwen-VL-Chat, the problem it generates, there is a discrepancy between the idea of the problem and the content of the problem, and the generated problem has not been tested for data rationality and does not conform to logic (it is paradoxical that there are two angles of 40$^{\circ}$ and $65^{\circ}$ in a right triangle).

On the other hand, Yi-VL-34B correctly interprets the meaning of the seed problem, analyzes the knowledge points tested, and modifies the problem scenario by adding conditions. However, the quality of its problem can be further improved because it introduces an invalid condition (the length of BC is meaningless for solving $\angle B$). Although there is no logical issue with this constructed problem, compared to our model's response, the problem constructed by our model is more reasonable.

\section{Conclusion}
In this work, we propose COMET, a ``Cone of Experience'' enhanced large multimodal model for mathematical problem generation. Inspired by the ``Cone of Experience'' theory, we follow the growth process of teachers to define the experience as symbolic, iconic, and direct. Based on this, we design a three-stage fine-tuning framework to enhance the capabilities of problem generation and problem solving within a single LMM to meet the requirements of educational applications. Moreover, a Chinese multimodal mathematics problem dataset (CMM12K) is built to alleviate the scarcity of Chinese multimodal corpora in this field. Extensive experiments have demonstrated the advancement and effectiveness of the proposed model. In the future, we will explore retrieval-enhanced generation methods based on model recall behavior, since the proposed direct experience can potentially serve as LMM historical memory.

\subsubsection*{Acknowledgments}
This work was financially supported by the National Science and Technology Major Project (Grant Nos. 2022ZD0117105), National Natural Science Foundation of China (Grant Nos. 62293554 and 62307015), China Postdoctoral Science Foundation (Grant Nos. 2023M741304 and 2023T160256), Hubei Provincial Natural Science Foundation of China (Grant Nos. 2023AFA020 and 2023AFB295), Fundamental Research Funds for the Central Universities (Grant Nos. CCNU23XJ007) and Knowledge Innovation Program of Wuhan-Shuguang Project (Grant Nos. 2023010201020390).

\bibliography{anthology,custom}
\bibliographystyle{acl_natbib}

% \appendix
% \section{Example Appendix}\label{sec:appendix}
% This is a section in the appendix.

\end{document}